\documentclass[10pt]{article}

\usepackage[utf8]{inputenc}
\usepackage[T1]{fontenc}
\usepackage{lmodern}
\usepackage[margin=1in,top=1in,bottom=1in]{geometry}
\usepackage{microtype}
\usepackage{amsmath,amssymb,amsthm,bm}
\usepackage{booktabs}
\usepackage{graphicx}
\usepackage{xcolor}
\usepackage[numbers,sort&compress,square,comma]{natbib}
\usepackage{hyperref}
\usepackage{cleveref}
\usepackage{placeins}

\usepackage{enumitem}
\usepackage{caption}
\usepackage{titlesec}
\usepackage{authblk}
\usepackage{algorithm}
\usepackage{algpseudocode}
\usepackage{tikz}
\usetikzlibrary{positioning,arrows.meta,shapes.geometric,calc,fit,backgrounds}

\hypersetup{
  colorlinks=true,
  linkcolor=blue!50!black,
  citecolor=blue!50!black,
  urlcolor=blue!70!black,
}

\titleformat*{\section}{\large\bfseries}
\titleformat*{\subsection}{\normalsize\bfseries}

\newcommand{\wvec}{\mathbf{w}}                
\newcommand{\fvec}{\mathbf{f}}                
\newcommand{\keepfrac}{\kappa}                
\newcommand{\valence}{V_{\!\text{aff}}}       
\newcommand{\arousal}{A}
\newcommand{\Rtask}{\mathcal{R}}              

\title{\textbf{Learning What to Remember:\\
A Cognitively Grounded Multi-Factor Value Model\\
for Agentic Memory}}
\author[1]{Zhibao Chen\thanks{Corresponding author: \texttt{maxzhibao@gmail.com}}}
\author[2]{Qian Cheng}
\affil[1]{Huatai Securities, Nanjing, China}
\affil[2]{OneBeget.com}
\date{}

\begin{document}
\maketitle

\begin{abstract}
\noindent
Long-running LLM agents accumulate interaction histories far larger than
any context window, forcing a standing decision: what to encode deeply,
what to forget, and what to retrieve under a fixed memory budget.
Production memory systems answer with \emph{semantic similarity} to the
current query or \emph{recency}. We argue both are mis-specified for the
governing decision, because the forgetting decision is made at
consolidation time, \emph{before} the future query is known.
We propose a \textbf{multi-factor memory value function}
$V(m)=\sum_i w_i f_i(m)$ over seven interpretable factors --- emotional
intensity, goal relevance, value alignment, self/user relevance, task
utility, reliability, and usage history --- each a determinant of human
retention drawn from cognitive psychology (value-directed remembering,
levels of processing, adaptive memory, emotional consolidation), whose
weights are \textbf{learned} from a downstream objective (gold-evidence
retention here; task-QA return in general) by a gradient-free optimiser,
and whose single scalar uniformly controls encoding depth, forget risk,
and retrieval rank.
We make a methodological point that reframes how memory retention is
measured: on the LongMemEval benchmark, scoring goal relevance against
the held-out evaluation question (an \emph{oracle} that peeks at the
future query) saturates gold-evidence retention at $\approx\!0.98$ for
similarity alone --- this measures \emph{retrieval}, not forgetting.
In the realistic \emph{blind} regime, where the consolidation policy
never sees the evaluation question, a learned multi-factor value retains
$\mathbf{0.770 \pm 0.011}$ of gold evidence across all $479$ usable
cases, versus $0.657$ for uniform weights, $0.518$ for the best single
factor, and $0.368$ for recency --- a per-case bootstrap places every
paired gap's $95\%$ confidence interval above zero, and a neural network
over the same factors ties the linear model, so the interpretable value
is not a compromise.
The learned blind weights are interpretable --- reliability, emotional
intensity, and self/user relevance dominate, while query-time goal
similarity is correctly down-weighted for the forgetting decision.
A controlled synthetic task with planted confounds confirms the learner
recovers a separating weighting ($1.00$ retention) where uniform
weighting trusts the confounds and fails ($0.62$).
The substrate is open-source; all reported experiments run on a single
CPU with no API calls.
\end{abstract}

\section{Introduction}
\label{sec:intro}

An LLM agent deployed over days or weeks accumulates an interaction
history far larger than any context window. It cannot attend to
everything, so on every consolidation step it makes a triage decision:
which observations to \emph{encode} (and how deeply), which to
\emph{forget}, and which to \emph{retrieve} when a new task arrives.
This decision is not cosmetic --- it determines whether the agent
remembers a user's stated preference three weeks later, whether it
recalls the one reliable fact among a hundred distractors, and whether
its working memory is dominated by recent chatter or by durable,
high-value knowledge.

Production memory stacks answer this triage with one of two
single-factor policies. Retrieval-augmented systems rank stored items by
\emph{semantic similarity} to the current query
\citep{lewis2020rag,packer2023memgpt}; time-to-live and sliding-window
schemes rank by \emph{recency}. Both are intuitive and both are
mis-specified for the governing decision. Similarity-to-query is only
defined \emph{at retrieval time}, once the query exists --- but the
\emph{forgetting} decision is made earlier, at consolidation, when the
future query is unknown. Recency is query-agnostic but discards the
common case of an old-but-important fact (a user's allergy, a project
constraint) in favour of recent low-value chatter. Neither policy has a
notion of how \emph{useful} a memory is to future behaviour.

Human memory is not single-factor. Decades of work show that what
survives forgetting is governed by the \emph{value} of an item:
people preferentially retain information tagged as high-value even at
the expense of low-value detail \citep{castel2008value}; depth of
processing --- semantic, self-referential, and goal-relevant encoding
--- predicts durability over shallow perceptual encoding
\citep{craik1972lop,rogers1977sre}; emotional arousal modulates
consolidation \citep{mcgaugh2000consolidation}; and the very shape of
forgetting tracks the \emph{need-probability} of information in the
environment, an adaptive rather than passive decay
\citep{anderson1991adaptive,ebbinghaus1885forgetting}. The common thread
is that retention is driven by a multi-factor estimate of an item's
expected future usefulness, not by any single cue.

We bring this principle to agentic memory. We define a
\textbf{multi-factor memory value function}
\begin{equation}
V(m) \;=\; \sum_{i} w_i\, f_i(m),
\label{eq:value-intro}
\end{equation}
a weighted combination of seven interpretable factors --- emotional
intensity, goal relevance, value alignment, self/user relevance, task
utility, reliability, and usage history. A \emph{single} scalar $V(m)$
then uniformly controls all three triage operations: encoding depth,
forget risk, and retrieval rank. Crucially, the weights $w_i$ are not
hand-set: they are \textbf{learned} from a downstream objective by a
gradient-free optimiser, because the
encode\,$\rightarrow$\,\allowbreak forget\,$\rightarrow$\,\allowbreak
retrieve\,$\rightarrow$\,\allowbreak answer
pipeline is non-differentiable. That objective is a stand-in for task
return; here we use gold-evidence retention under a fixed memory budget
(an API-free proxy), while a fully instrumented setting would use
downstream QA accuracy. The value function is thus fit to maximise the
chosen objective, in the spirit of expected-utility credit assignment
\citep{sutton2018rl}.

Along the way we surface a methodological pitfall in how memory
retention is commonly measured, and turn it into a clean experimental
contrast. On the LongMemEval benchmark \citep{wu2025longmemeval}, the
``gold'' evidence for a question is, by construction, defined
\emph{relative to that question}. If a memory policy is allowed to score
goal relevance as the cosine similarity between a stored turn and the
held-out evaluation question --- an \emph{oracle} that peeks at the
future query --- then similarity alone retains essentially all gold
evidence ($\approx\!0.98$), and no multi-factor advantage is possible.
But this measures \emph{retrieval}, not forgetting: a real agent has not
seen the evaluation question at consolidation time. In the realistic
\emph{blind} regime, where goal relevance is computed only against
information available at consolidation (the topic of the ongoing
session), the picture inverts: similarity collapses to near-chance and a
learned multi-factor value retains gold far better than any single-factor
baseline. Separating these two regimes is, we argue, necessary for any
honest evaluation of a forgetting policy.

\paragraph{Contributions.}
\begin{enumerate}[leftmargin=1.4em,itemsep=2pt,topsep=2pt]
\item A \textbf{multi-factor memory value function}
(\cref{eq:value-intro}) for the expected usefulness of a memory to
future agent behaviour, with seven interpretable factors, and a single
scalar that uniformly drives encoding depth, forgetting, and retrieval
(\cref{sec:method}).
\item A \textbf{learned-weight} formulation (A2): a gradient-free
optimiser fits $\wvec$ to a downstream objective (gold-evidence retention
here; task-QA return in general), replacing hand-tuned resistances with
credit-assigned weights (\cref{sec:method:learn}).
\item A \textbf{methodological contrast} --- \emph{oracle} vs.\
\emph{blind} forgetting --- that separates retrieval from retention and
shows why query-defined retention benchmarks cannot, on their own,
demonstrate a forgetting advantage (\cref{sec:exp:blind}).
\item \textbf{Empirical results} on all $479$ usable LongMemEval-S cases:
in the blind regime, learned multi-factor value retains $0.770\pm0.011$ of
gold evidence versus $0.657$ (uniform), $0.518$ (best single factor), and
$0.368$ (recency), with every per-case bootstrap gap above zero; a neural
MLP over the same factors ties the linear model ($+0.003$), so the
interpretable linear value suffices (\cref{sec:exp:blind}). A synthetic
confound study validates the learner (\cref{sec:exp:synth}).
\item An \textbf{open-source, CPU-only} substrate and evaluation harness;
every reported number reproduces with no API calls.
\end{enumerate}

\FloatBarrier
\section{Related Work}
\label{sec:related}

\paragraph{Memory systems for LLM agents.}
Retrieval-augmented generation retrieves passages by embedding similarity
to the query \citep{lewis2020rag}, and most agent memory frameworks
inherit this ranking. MemGPT \citep{packer2023memgpt} treats context as a
paged virtual memory with explicit eviction, but eviction is driven by
recency and capacity pressure rather than a learned value. Generative
Agents \citep{park2023generative} introduce a retrieval score that blends
recency, importance, and relevance --- importantly, a \emph{multi-factor}
score --- but the importance term is produced ad hoc by an LLM rating
$1$--$10$, the weights are fixed and hand-set, and the score governs only
retrieval, not encoding depth or forgetting. MemoryBank
\citep{zhong2024memorybank} adds an Ebbinghaus-inspired decay so older
memories fade unless reinforced, again with hand-set dynamics. MemOS
\citep{memos2025} frames memory as an operating-system resource with
scheduling and lifecycle management, motivating a principled value signal
but not learning one; the broader cognitive-architecture view of language
agents \citep{sumers2024cogarch} similarly treats memory as a managed
module without prescribing how its contents are valued. Across this line
of work, the triage signal is typically (i) single-factor (similarity or
recency), or (ii) multi-factor but hand-weighted and confined to
retrieval. We differ on
both counts: our value is multi-factor, its weights are \emph{learned}
from a downstream objective, and the \emph{same} scalar drives encoding,
forgetting, and retrieval.

\paragraph{Value- and need-based retention in cognition.}
The factors we combine are not arbitrary; each corresponds to a robust
determinant of human retention. Value-directed remembering shows people
strategically retain high-value items and shed low-value detail
\citep{castel2008value}. Levels-of-processing
\citep{craik1972lop} and the self-reference effect \citep{rogers1977sre}
establish that semantically, goal-, and self-relevant encoding is more
durable than shallow encoding. Emotional arousal modulates consolidation
\citep{mcgaugh2000consolidation}. The rational analysis of memory
\citep{anderson1991adaptive} reframes forgetting itself as adaptive: the
retention function mirrors the environmental \emph{need-probability} of
information. Relatedly, the metamemory and desirable-difficulties
tradition argues that forgetting and selective retention serve memory
rather than merely degrade it \citep{bjork1994memory}. Our contribution
is to operationalise this
multi-determinant view as a single learned scalar for an artificial
agent, rather than to model human data --- a companion paper treats the
cognitive-modeling angle; here the target is engineering utility.

\paragraph{Learning what to remember.}
Reinforcement learning provides the natural framing for fitting a memory
policy to downstream outcome: the value of retaining an item is its
marginal contribution to future return \citep{sutton2018rl}. Because the
encode--forget--retrieve--answer pipeline is non-differentiable
(discrete keep/drop decisions, an external answerer), we fit weights with
a gradient-free black-box optimiser \citep{hansen2016cmaes} rather than
backpropagation. This keeps the value function low-dimensional (seven
weights) and interpretable, in contrast to end-to-end learned memory
controllers whose policies are opaque.

\paragraph{Evaluating memory.}
LongMemEval \citep{wu2025longmemeval} is the most directly relevant
benchmark: $500$ questions over long multi-session chat histories, with
gold-evidence turns flagged inside a haystack of distractor sessions. We
use it as our real-data testbed. Our methodological observation ---
that scoring relevance against the held-out question conflates retrieval
with retention, and that a forgetting policy must be evaluated
\emph{blind} to the future query --- applies to any retention metric
built on query-defined gold, and to our knowledge has not been made
explicit in this setting.

\FloatBarrier
\section{Method}
\label{sec:method}

\subsection{A multi-factor memory value}
\label{sec:method:value}

Each memory $m$ (a stored turn or consolidated item) is summarised by a
factor vector $\fvec(m)\in[0,1]^7$ and scored by a learned linear value
\begin{equation}
V(m) \;=\; \sum_{i=1}^{7} w_i\, f_i(m),
\qquad \wvec \in \mathbb{R}_{\ge 0}^{7}.
\label{eq:value}
\end{equation}
The seven factors are chosen so that each is an interpretable,
independently computable determinant of expected future usefulness
(\cref{tab:factors}). We regard \cref{eq:value} as a \emph{learned reward
model for memory}: $V(m)$ estimates a memory's contribution to future
task return, and its weights are fit by policy optimisation
(\cref{sec:method:learn}) rather than set by hand.

Linearity here is a deliberate design choice, not a limitation we
tolerate, for three reasons. \textbf{(i) Capacity match}: with seven
interpretable factors and limited, indirect supervision, a
seven-parameter value is the appropriate capacity; richer function
classes overfit this regime (\cref{sec:limitations}). \textbf{(ii)
Auditability}: the learned weights $w_i$ \emph{are} the explanation --- a
deployment can read off which factors a given workload rewards
(\cref{sec:exp:blind}), which a black-box scorer cannot provide and which
is a practical requirement for a memory controller. \textbf{(iii)
Decision-theoretic reading}: \cref{eq:value} is the first-order
(additive-utility) approximation of a memory's expected contribution to
return --- the principled starting point before modelling factor
interactions. More generally the value is any learned scoring function
$V(m)=g_\theta(\fvec(m))$, and \cref{eq:value} is its interpretable linear
instance; a neural $g_\theta$ is validated as an \emph{interaction
ablation} in \cref{sec:exp:blind} --- an MLP over the same factors ties
the linear model, confirming they combine near-additively, so the linear
value stays the default.

\begin{table}[ht!]
\centering\small
\caption{The seven memory-value factors. The right column marks which
factors are populated by the \emph{API-free} annotator used in our
experiments; the remaining three require a value profile, an LLM judge,
or access logs and are held at $0$ here (\cref{sec:limitations}).}
\label{tab:factors}
\begin{tabular}{llc}
\toprule
Factor & Definition (API-free form) & Live here? \\
\midrule
emotional intensity & $|\valence|\cdot\arousal$ from a lexical/structural extractor & yes \\
goal relevance      & $\tfrac12(1+\cos(\mathbf{e}_m, \mathbf{e}_{\text{goal}}))$ & yes \\
self/user relevance & $\tfrac12(1+\cos(\mathbf{e}_m, \mu_{\text{user}}))$ & yes \\
reliability         & provenance heuristic (user-stated $>$ model-stated) & yes \\
value alignment     & match to a value profile (SOUL) & no ($0$) \\
task utility        & marginal task-success contribution (LLM judge) & no ($0$) \\
usage history       & $r/(1+r)$, $r=$ retrieval count & no ($0$) \\
\bottomrule
\end{tabular}
\end{table}

Here $\mathbf{e}_m$ is a sentence embedding of the memory text
(all-MiniLM-L6-v2, $384$-dim; \citealp{reimers2019sbert}),
$\mu_{\text{user}}$ is the running centroid of the user's turns (a
query-agnostic ``what this user is about'' anchor), and
$\mathbf{e}_{\text{goal}}$ is the embedding of the agent's active goal.
The goal anchor is the pivot of our central experiment
(\cref{sec:exp:blind}): at retrieval time it is the query, but at
\emph{consolidation} time the agent must use a query-agnostic proxy.

\subsection{One value, three operations}
\label{sec:method:control}

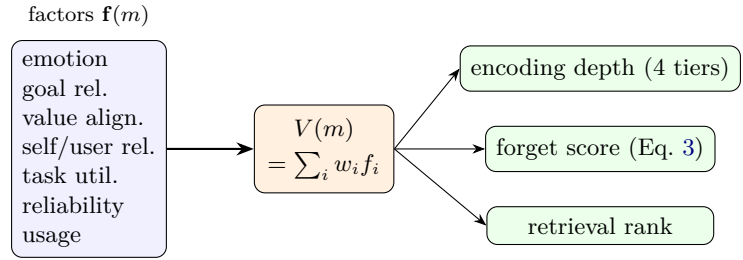
\begin{figure}[ht!]
\centering
\begin{tikzpicture}[
  font=\small, >=Stealth, node distance=0.45cm and 1.15cm,
  box/.style={draw, rounded corners, align=left, inner sep=4pt},
  val/.style={draw, rounded corners, fill=orange!12, align=center,
              inner sep=5pt, minimum height=1.0cm},
  op/.style={draw, rounded corners, fill=green!8, align=center,
             inner sep=4pt, minimum width=3.0cm}]
\node[box, fill=blue!6] (f)
  {emotion\\ goal rel.\\ value align.\\ self/user rel.\\
   task util.\\ reliability\\ usage};
\node[above=0.05cm of f, font=\footnotesize] {factors $\fvec(m)$};
\node[val, right=of f] (v) {$V(m)$\\[2pt]$=\sum_i w_i f_i$};
\node[op, right=1.2cm of v] (forget) {forget score (Eq.~\ref{eq:forget})};
\node[op, above=of forget] (enc) {encoding depth (4 tiers)};
\node[op, below=of forget] (ret) {retrieval rank};
\draw[->, thick] (f) -- (v);
\draw[->] (v.east) -- (enc.west);
\draw[->] (v.east) -- (forget.west);
\draw[->] (v.east) -- (ret.west);
\end{tikzpicture}
\caption{\textbf{One value, three operations.} A single learned scalar
$V(m)=\sum_i w_i f_i(m)$ over seven factors drives encoding depth,
forgetting, and retrieval, replacing three separately tuned rules. The
weights $w_i$ are fit to a downstream objective (\cref{sec:method:learn}).}
\label{fig:schematic}
\end{figure}

The single scalar $V(m)$ drives all three memory operations
(\cref{fig:schematic}), replacing the separate, hand-tuned rules of prior
systems.

\paragraph{Encoding depth.}
Following levels-of-processing \citep{craik1972lop}, higher-value items
are encoded more deeply. We map the normalised value
$\bar V(m)=V(m)/\sum_i w_i \in[0,1]$ to four encoding tiers
(shallow / semantic / schematic / meta) by thresholds
$\{0.20,0.45,0.70\}$, so deeper encoding is allocated to higher-value
memories at write time.

\paragraph{Forgetting.}
The forget score combines universal time/access dynamics with a
value-driven resistance:
\begin{equation}
\mathrm{forget}(m) \;=\;
\underbrace{(\Delta t)^{0.7}}_{\text{recency}}
\cdot
\underbrace{\tfrac{1}{1+r}}_{\text{usage}}
\cdot
\underbrace{\tfrac{1}{1+\beta\,V(m)}}_{\text{value resistance}},
\label{eq:forget}
\end{equation}
where $\Delta t$ is age in days and $r$ is retrieval count. High value
$V(m)$ multiplicatively resists forgetting, generalising the
emotion-only resistance of prior cognitive memory models to the full
factor set. Items with the highest forget score are dropped first when
the store exceeds budget.

\paragraph{Retrieval.}
At query time the same value contributes to the retrieval rank,
alongside query-specific similarity and mood congruence, so that durable
high-value memories are surfaced preferentially. Because our central
evaluation isolates the budgeted keep/drop decision
(\cref{sec:exp:blind}), we rank candidates directly by $V(m)$ --- using
the regime-specific goal anchor defined there (the question under
\emph{oracle}, the session topic under \emph{blind}) --- and report
retention of the gold set.

\subsection{Learning the weights from task return}
\label{sec:method:learn}

The weights $\wvec$ are \emph{learned}, not hand-set. Let a memory policy
parameterised by $\wvec$ run the
encode\,$\rightarrow$\,\allowbreak forget\,$\rightarrow$\,\allowbreak
retrieve\,$\rightarrow$\,\allowbreak answer
pipeline on a set of training episodes and return a scalar
$\Rtask(\wvec)$ (here, mean gold-evidence retention under a fixed keep
budget; in a fully API-gated setting, downstream QA accuracy). Because
the pipeline contains discrete keep/drop decisions and an external
answerer, $\Rtask$ is non-differentiable in $\wvec$. We therefore
maximise it with a gradient-free stochastic hill-climb
(\cref{alg:learn}), a lightweight stand-in for CMA-ES
\citep{hansen2016cmaes} appropriate to a seven-dimensional search.

\begin{algorithm}[ht!]
\small
\caption{Gradient-free weight learning (A2).}
\label{alg:learn}
\begin{algorithmic}[1]
\State $\wvec \gets \mathbf{1}$ \Comment{uniform init}
\State $R^\star \gets \Rtask(\wvec)$; \; $\sigma \gets \sigma_0$
\For{$t = 1 \dots T$}
  \State $\wvec' \gets \wvec$
  \For{each factor $i$}
    \If{$\mathrm{rand}() < 0.5$}
      $\wvec'_i \gets \max(0,\; \wvec_i + \mathcal{N}(0,\sigma))$
    \EndIf
  \EndFor
  \State $R' \gets \Rtask(\wvec')$
  \If{$R' > R^\star$} $\wvec \gets \wvec'$; \; $R^\star \gets R'$ \EndIf
  \State $\sigma \gets \gamma\,\sigma$ \Comment{anneal}
\EndFor
\State \Return $\wvec$
\end{algorithmic}
\end{algorithm}

The objective is evaluated on a held-in training split and the learned
$\wvec$ is reported on a disjoint test split; best-so-far return is
monotone non-decreasing by construction. To keep the learned weights
honest and interpretable, we restrict the search to the factors actually
populated by the annotator in a given setting --- otherwise the
optimiser assigns arbitrary non-zero weight to factors that are
identically $0$ and therefore multiply out of \cref{eq:value}.

\FloatBarrier
\section{Experiments}
\label{sec:exp}

We ask two questions. First, can the gradient-free learner actually
recover a useful weighting when factors are confounded
(\cref{sec:exp:synth})? Second --- the headline --- does a learned
multi-factor value retain gold evidence better than single-factor
baselines on a \emph{real} long-horizon benchmark, once we evaluate
forgetting honestly (\cref{sec:exp:blind})? Both experiments are
CPU-only and use no API calls; embeddings come from a local
sentence-transformer \citep{reimers2019sbert}.

\subsection{Synthetic confound study: does the learner work?}
\label{sec:exp:synth}

\paragraph{Design.}
Each synthetic case contains $4$ gold items and $16$ distractors. Gold
items are \emph{old} but carry the true signal (high goal relevance,
task utility, reliability); distractors are \emph{recent} and carry two
deliberately planted confounds (high value alignment and self relevance)
plus high emotional noise. A policy that trusts all factors equally is
misled by the confounds; a recency- or emotion-only policy keeps the
distractors. We learn $\wvec$ on $60$ training cases to maximise
gold retention at keep-fraction $\keepfrac=0.4$, and report retention on
$60$ disjoint test cases.

\paragraph{Result.}
\Cref{tab:synth} shows that the learner recovers a weighting that retains
\emph{all} gold ($1.00$) on held-out cases, while uniform weighting ---
which trusts the confounds equally --- retains only $0.62$, and
the emotion-, self-, and recency-only policies we test retain $0.00$
(they rank the recent, emotionally noisy distractors first). The learned
weights place the largest mass on the true-signal factors (reliability
$1.36$, task utility $0.88$, goal relevance $0.49$) and drive the
\emph{value-alignment} confound down to $0.09$; the self-relevance
confound is \emph{not} fully suppressed ($1.01$), but the true-signal mass
alone suffices to separate gold from distractors. This is a sanity check,
not the central claim: it shows only that the seven-dimensional black-box
search can fit a \emph{separating mixture} where uniform weighting --- and
the single-factor policies we test --- cannot. We do not test goal-,
task-, or reliability-only baselines on the synthetic case; the real-data
study below tests all four live single factors directly.

\begin{table}[ht!]
\centering\small
\caption{Synthetic confound study: gold retention on $60$ held-out cases
(keep $40\%$). Gold is signalled by goal/task/reliability; distractors
carry planted value-alignment and self-relevance confounds.}
\label{tab:synth}
\begin{tabular}{lr}
\toprule
Policy & Gold retention \\
\midrule
\textbf{learned\_V} & $\mathbf{1.00}$ \\
uniform\_V          & $0.62$ \\
emotion\_only       & $0.00$ \\
self\_only          & $0.00$ \\
recency\_only       & $0.00$ \\
\bottomrule
\end{tabular}
\end{table}

\subsection{Real LongMemEval: blind vs.\ oracle forgetting}
\label{sec:exp:blind}

\paragraph{Setup.}
We use LongMemEval-S \citep{wu2025longmemeval}: questions over long
multi-session chat histories ($\sim\!500$ turns per case) with
gold-evidence turns flagged inside a haystack of distractor sessions. Of
the $500$ questions, $479$ carry at least one flagged gold turn; we use
all $479$. For each case we annotate every turn with the four API-free
factors (emotional intensity, goal relevance, self/user relevance,
reliability; \cref{tab:factors}), rank turns by $V(m)$, keep the top
$\keepfrac=30\%$, and measure the fraction of flagged gold turns
retained. Weights are learned over the live factors on a training split
and evaluated on a disjoint test split; we report mean $\pm$ standard
deviation over $20$ resampled $50/50$ splits ($240$ test cases per rep),
and confirm significance with a per-case bootstrap below. This isolates
the budgeted keep/drop (retention) decision: we do not run an answerer or
the full encode/forget/retrieve loop here, and we report gold-evidence
retention rather than QA accuracy (\cref{sec:limitations}). To pre-compute
the factor annotations once and reuse them across all configurations, we
cache the per-turn factors; every number below is API-free and reproduces
on a single CPU.

\paragraph{The oracle/blind distinction.}
Gold in LongMemEval is defined relative to the evaluation question, so
the \emph{goal-relevance} factor can be computed two ways:
\begin{itemize}[leftmargin=1.4em,itemsep=1pt,topsep=2pt]
\item \textbf{Oracle:} $\mathbf{e}_{\text{goal}}$ is the embedding of the
held-out evaluation question. This peeks at the future query --- a
\emph{retrieval} signal, unavailable when the agent actually forgets.
\item \textbf{Blind:} $\mathbf{e}_{\text{goal}}$ is the centroid of the
ongoing session's user turns (``what is this session about''), the
information genuinely available at consolidation time.
\end{itemize}
All other factors are identical across regimes; only the goal anchor
moves. This is exactly the difference between measuring retrieval and
measuring forgetting.

\begin{figure}[ht!]
\centering
\includegraphics[width=0.82\linewidth]{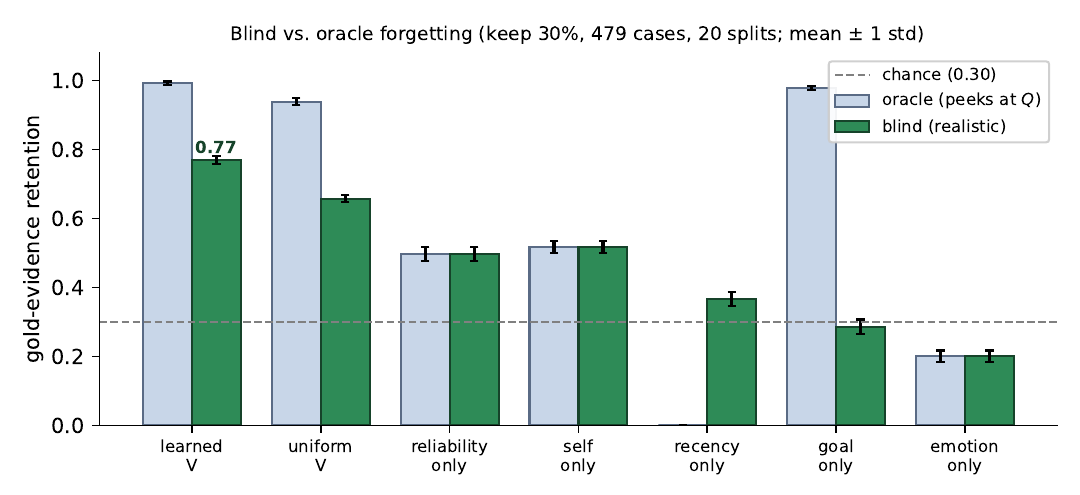}
\caption{\textbf{Blind vs.\ oracle gold retention} (keep $30\%$; all $479$
usable LongMemEval-S cases; $20$ resampled splits, mean $\pm 1$ std). Under
the \emph{oracle} goal anchor (cosine to the held-out question), similarity
alone saturates retention --- a retrieval ceiling. Under the realistic
\emph{blind} anchor (session topic only), every single-factor and recency
baseline collapses toward the chance floor, while the learned multi-factor
value retains $0.77$. Recency has no oracle variant.}
\label{fig:blind}
\end{figure}

\paragraph{Result.}
\Cref{tab:blind} and \cref{fig:blind} report both regimes. In the \textbf{oracle} regime,
goal relevance alone retains $0.979$ of gold and the learned value reaches
$0.993$ (uniform $0.939$) --- all near the ceiling: once you can score
similarity to the question, similarity is all you need, and there is no
room for a multi-factor advantage. This is the honest negative that a
query-defined retention metric produces, and it is why such metrics
cannot, by themselves, evaluate a forgetting policy.

In the realistic \textbf{blind} regime the ordering inverts.
Goal-relevance-to-session collapses to $0.286$ (near the $0.30$ chance
floor), emotion to $0.201$, recency to $0.368$, reliability to $0.497$,
and self-relevance --- the strongest single factor --- to $0.518$: no
single cue identifies the future needle without the question. The
\emph{learned multi-factor value} retains $\mathbf{0.770\pm0.011}$,
beating uniform weighting ($0.657$), the best single factor (self,
$0.518$), and recency ($0.368$). A per-case bootstrap ($B{=}10^3$,
resampling the $479$ cases) puts the paired gaps at $+0.120$ vs.\ uniform
($95\%$ CI $[0.091,0.149]$), $+0.402$ vs.\ recency ($[0.350,0.452]$), and
$+0.277$ vs.\ reliability-only ($[0.235,0.319]$); every interval excludes
zero, and the learned policy wins on all $20$ resampled splits.
Multi-factor value's advantage is thus precisely where it should be: in
the query-agnostic forgetting decision, not in query-defined retrieval.

\begin{table}[ht!]
\centering\small
\caption{LongMemEval gold retention (keep $30\%$; all $479$ usable cases;
mean $\pm$ std over $20$ resampled test splits, $240$ cases each).
\emph{Oracle} scores goal relevance against the held-out question
(retrieval ceiling); \emph{blind} scores it against the session topic
(realistic forgetting). Self-, emotion-, and reliability-only are
regime-invariant by construction (only the goal anchor moves).}
\label{tab:blind}
\begin{tabular}{lcc}
\toprule
Policy & Oracle (peeks at $Q$) & Blind (realistic) \\
\midrule
\textbf{learned\_V}  & $0.993 \pm 0.005$ & $\mathbf{0.770 \pm 0.011}$ \\
uniform\_V           & $0.939 \pm 0.011$ & $0.657 \pm 0.011$ \\
self\_only           & $0.518 \pm 0.017$ & $0.518 \pm 0.017$ \\
reliability\_only    & $0.497 \pm 0.020$ & $0.497 \pm 0.020$ \\
goal\_only           & $0.979 \pm 0.006$ & $0.286 \pm 0.021$ \\
emotion\_only        & $0.201 \pm 0.017$ & $0.201 \pm 0.017$ \\
recency\_only        & ---               & $0.368 \pm 0.021$ \\
random (keep $30\%$) & ---               & $0.300$ \\
\bottomrule
\end{tabular}
\end{table}

\paragraph{Learned weights are interpretable.}
The mean learned blind weights are reliability $0.64$, emotional
intensity $0.55$, self/user relevance $0.23$, and goal relevance $0.00$.
The policy discovered, from retention alone, that LongMemEval gold is
dominated by \emph{reliable, user-stated, often self-relevant} facts ---
and that session-topic similarity (the blind goal anchor) is useless for
predicting the future needle, so it is driven to zero. This is the
value-directed signal a recency or similarity policy structurally cannot
represent, and it accounts for the $+0.11$ to $+0.40$ retention gap.
(Self-relevance is the best \emph{standalone} factor in \cref{tab:blind}
yet reliability carries the larger learned weight: the weights measure
each factor's marginal contribution \emph{within} the learned mixture, not
its solo ranking power.)

\paragraph{Does a neural value help? A nonlinear ablation.}
Our value is linear by design (\cref{sec:method:value}); a natural worry
is that it leaves performance on the table by ignoring factor
interactions (e.g.\ emotion mattering only for self-relevant turns). We
test this directly: we replace the linear value with a small MLP
$g_\theta$ over the same four live factors ($[16,16]$ hidden units),
trained by a pairwise logistic ranking loss (gold ranked above non-gold
within each case), and evaluate it under the identical blind protocol.
The MLP retains $0.773\pm0.011$ versus the linear model's
$0.770\pm0.011$ --- a paired difference of $+0.003\pm0.013$, winning on
only $55\%$ of splits, i.e.\ a statistical tie. On these four live factors
the nonlinear capacity buys nothing: the factors combine near-additively,
so the interpretable linear model is not a compromise but the right choice.
The MLP is an ablation, not a replacement.

\begin{figure}[ht!]
\centering
\includegraphics[width=0.74\linewidth]{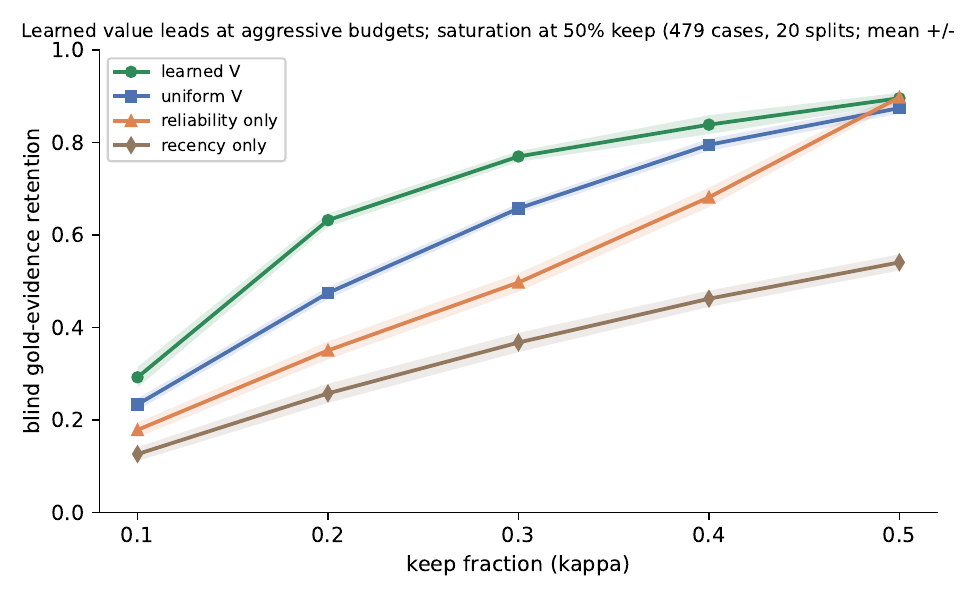}
\caption{\textbf{Keep-fraction sweep} (blind regime, $479$ cases, $20$
splits, mean $\pm 1$ std). The learned multi-factor value dominates uniform,
reliability-only, and recency at every aggressive budget
($\keepfrac \le 0.4$); the gap closes only at the near-trivial
$\keepfrac{=}0.5$ where the benchmark saturates.}
\label{fig:keepfrac}
\end{figure}

\paragraph{The advantage is not budget-specific.}
\Cref{fig:keepfrac} sweeps the keep fraction $\keepfrac \in
\{0.1,\dots,0.5\}$. The learned value leads at every realistic forgetting
budget ($\keepfrac \le 0.4$); only at $\keepfrac{=}0.5$ --- a near-trivial
$50\%$ keep, where all factor-based policies saturate at $0.87$--$0.90$
--- does the best single factor draw level. The multi-factor advantage is
thus largest exactly where forgetting is aggressive and the decision
matters most.

\FloatBarrier
\section{Discussion}
\label{sec:discussion}

\paragraph{One scalar, three operations.}
The practical appeal of \cref{eq:value} is consolidation: a single
learned value replaces the separate, independently tuned rules that
production systems use for write-time importance, eviction, and
retrieval ranking. A deployment exposes one seven-dimensional knob, and
\cref{eq:forget} and the encoding tiers inherit any change to it
automatically. Because the value is linear, the learned weights are a
\emph{readable audit} of the policy: \cref{sec:exp:blind} shows the
fitted policy says, in plain terms, ``keep reliable user-stated facts;
do not trust session-topic similarity for forgetting.''

\paragraph{The oracle/blind lesson generalises.}
Our sharpest methodological point is independent of the value function.
Any retention metric whose gold set is defined relative to a future
query will reward a policy that scores relevance \emph{to that query}
--- but that policy is unavailable at forgetting time. Reporting such
numbers measures an upper bound on retrieval, not the quality of
forgetting. Evaluations of agent forgetting should therefore (i) compute
relevance only from information available at consolidation, and (ii)
report the oracle ceiling alongside, so the retrieval/retention gap is
visible. The $0.979 \rightarrow 0.286$ collapse of goal-only between our
two regimes (\cref{tab:blind}) is the size of the artefact a careless
metric would hide.

\paragraph{Learned weights as a workload diagnostic.}
Because the weights are fit to the chosen objective, they characterise the
workload, not just the policy. On LongMemEval the signal is reliability
and self-relevance; a coding agent's workload might instead reward task
utility and goal relevance. The same harness, pointed at a new return
signal, yields a new interpretable weighting --- a portable recipe for
\emph{discovering} what a given agent should remember, rather than
guessing it. This extends the hand-set recency/importance/relevance blend
of Generative Agents \citep{park2023generative} in two ways: the blend is
learned, and it governs encoding and forgetting, not only retrieval.

\FloatBarrier
\section{Limitations}
\label{sec:limitations}

We are deliberately explicit about what the present results do and do not
establish.

\paragraph{L1 --- Retention is a proxy for end-task accuracy.}
Our metric is gold-evidence retention under a keep budget, not downstream
QA correctness. High retention is \emph{necessary} for a correct answer
that depends on the evidence, but not \emph{sufficient}: the answerer
must also retrieve and use the retained turn. Measuring final QA accuracy
under each memory policy requires an answerer LLM and an LLM judge, which
we leave to an API-gated follow-up. The claim here is about what survives
forgetting, not about answer quality.

\paragraph{L2 --- Three of seven factors are inert in our annotator.}
Value alignment, task utility, and usage history are held at $0$ in the
API-free setting (\cref{tab:factors}): they need a value profile, an LLM
judge, and access logs respectively. The blind headline therefore uses
only four live factors, and task utility --- arguably the most direct
expected-utility signal --- is exactly the one an answerer would unlock.
Our result is thus a \emph{conservative, four-factor, API-free} estimate;
the three inert factors might raise performance or, if noisy, lower it, so
we do not claim the full function is strictly better --- only that a
useful value signal is already recoverable with no API budget.

\paragraph{L3 --- Single benchmark and modality.}
We evaluate on LongMemEval-S (all $479$ usable cases of the $500$),
English multi-session chat. Other agentic workloads --- tool use, coding,
web navigation --- have different gold structure and may favour different
factors; we do not yet test them. The synthetic study controls confounds
but is, by design, not naturalistic.

\paragraph{L4 --- The blind goal anchor is one choice among many.}
We proxy the consolidation-time goal by the session's user-turn centroid.
Other query-agnostic anchors (an explicit task description, a running
objective stack) could carry more signal and would change the blind
goal-relevance term. We claim the \emph{regime distinction} is essential,
not that the session centroid is optimal.

\paragraph{L5 --- Linear value and a simple optimiser.}
\Cref{eq:value} is linear. We tested whether this costs accuracy by
running a neural $g_\theta$ ablation (\cref{sec:exp:blind}): a small MLP
over the same factors, trained with a learning-to-rank loss, ties the
linear model ($+0.003\pm0.013$), so on these factors linearity is not a
compromise --- though a different factor set or workload could expose
interactions the linear form cannot represent (e.g.\ emotion mattering
only for self-relevant items). The learner is a random-restart hill-climb,
not CMA-ES; a stronger optimiser could shift the fitted weights, though
the blind result is stable across $20$ resampled splits and a per-case
bootstrap (\cref{sec:exp:blind}) places every key gap's $95\%$ CI above
zero. Because every downstream operation consumes the value only through
the scalar $V(m)$, swapping the scoring function --- linear or neural ---
leaves the encoding/forgetting/retrieval machinery unchanged.

\FloatBarrier
\section{Conclusion}
\label{sec:conclusion}

We argued that the standing triage decision in agentic memory --- what to
encode, forget, and retrieve under a fixed budget --- is mis-served by
the single-factor similarity and recency policies in common use, because
the forgetting decision is made before the future query is known. We
proposed a multi-factor memory value $V(m)=\sum_i w_i f_i(m)$ whose
weights are learned from a downstream objective (gold-evidence retention
here; task-QA return in general) and whose single scalar uniformly
governs all three operations. On all $479$ usable LongMemEval-S cases, a
learned multi-factor value retains $0.770\pm0.011$ of gold evidence in the
realistic blind-forgetting regime, versus $0.657$ for uniform weighting,
$0.518$ for the best single factor, and $0.368$ for recency --- every
per-case bootstrap gap excludes zero, and a neural MLP over the same
factors ties the linear model, so the interpretable linear value is not a
compromise. The learned weights recover the workload's true signal. We also showed why this advantage is invisible to a query-defined
retention metric: under an oracle that peeks at the evaluation question,
similarity alone saturates retention and the multi-factor gap vanishes
--- a distinction any honest evaluation of forgetting must draw.

The immediate next step is to close the loop to end-task accuracy:
fitting the weights to downstream QA correctness (unlocking the
task-utility factor) and measuring answer quality, not just retention,
under each policy. Beyond that, the same learned-value harness applied to
tool-use and coding agents would test whether the value-directed view of
memory transfers across agentic workloads. The substrate, factor
annotator, and evaluation harness are open-source
\citep{chen2026lmfm}, and every number in this paper reproduces on a
single CPU.

\FloatBarrier

\bibliographystyle{plainnat}
\bibliography{references}

\begin{thebibliography}{18}
\providecommand{\natexlab}[1]{#1}
\providecommand{\url}[1]{\texttt{#1}}
\expandafter\ifx\csname urlstyle\endcsname\relax
  \providecommand{\doi}[1]{doi: #1}\else
  \providecommand{\doi}{doi: \begingroup \urlstyle{rm}\Url}\fi

\bibitem[Anderson and Schooler(1991)]{anderson1991adaptive}
John~R. Anderson and Lael~J. Schooler.
\newblock Reflections of the environment in memory.
\newblock \emph{Psychological Science}, 2\penalty0 (6):\penalty0 396--408,
  1991.

\bibitem[Bjork(1994)]{bjork1994memory}
Robert~A. Bjork.
\newblock Memory and metamemory considerations in the training of human beings.
\newblock In Janet Metcalfe and Arthur~P. Shimamura, editors,
  \emph{Metacognition: Knowing about Knowing}, pages 185--205. MIT Press, 1994.

\bibitem[Castel(2007)]{castel2008value}
Alan~D. Castel.
\newblock The adaptive and strategic use of memory by older adults: Evaluative
  processing and value-directed remembering.
\newblock In Aaron~S. Benjamin and Brian~H. Ross, editors, \emph{Psychology of
  Learning and Motivation}, volume~48, pages 225--270. Academic Press, 2007.
\newblock \doi{10.1016/S0079-7421(07)48006-9}.

\bibitem[Chen(2026)]{chen2026lmfm}
Zhibao Chen.
\newblock {L}earning-{M}ulti-{F}actor-{M}emory: Open-source implementation of
  the multi-factor value model for agentic memory.
\newblock \url{https://github.com/zhibao-dev/Learning-Multi-Factor-Memory},
  2026.

\bibitem[Craik and Lockhart(1972)]{craik1972lop}
Fergus I.~M. Craik and Robert~S. Lockhart.
\newblock Levels of processing: A framework for memory research.
\newblock \emph{Journal of Verbal Learning and Verbal Behavior}, 11\penalty0
  (6):\penalty0 671--684, 1972.

\bibitem[Ebbinghaus(1885)]{ebbinghaus1885forgetting}
Hermann Ebbinghaus.
\newblock \emph{{\"U}ber das Ged{\"a}chtnis: Untersuchungen zur experimentellen
  Psychologie}.
\newblock Duncker \& Humblot, Leipzig, 1885.
\newblock English translation by Ruger \& Bussenius, Teachers College, 1913.

\bibitem[Hansen(2016)]{hansen2016cmaes}
Nikolaus Hansen.
\newblock The {CMA} evolution strategy: A tutorial.
\newblock \emph{arXiv preprint arXiv:1604.00772}, 2016.
\newblock Tutorial reference for gradient-free black-box optimisation.

\bibitem[Lewis et~al.(2020)Lewis, Perez, Piktus, Petroni, Karpukhin, Goyal,
  K{\"u}ttler, Lewis, Yih, Rockt{\"a}schel, Riedel, and Kiela]{lewis2020rag}
Patrick Lewis, Ethan Perez, Aleksandra Piktus, Fabio Petroni, Vladimir
  Karpukhin, Naman Goyal, Heinrich K{\"u}ttler, Mike Lewis, Wen-tau Yih, Tim
  Rockt{\"a}schel, Sebastian Riedel, and Douwe Kiela.
\newblock Retrieval-augmented generation for knowledge-intensive {NLP} tasks.
\newblock In \emph{Advances in Neural Information Processing Systems
  (NeurIPS)}, 2020.

\bibitem[Li et~al.(2025)Li, Xi, Li, Chen, Chen, Song, et~al.]{memos2025}
Zhiyu Li, Chenyang Xi, Chunyu Li, Ding Chen, Boyu Chen, Shichao Song, et~al.
\newblock {MemOS}: A memory {OS} for {AI} system, 2025.

\bibitem[McGaugh(2000)]{mcgaugh2000consolidation}
James~L. McGaugh.
\newblock Memory --- a century of consolidation.
\newblock \emph{Science}, 287\penalty0 (5451):\penalty0 248--251, 2000.

\bibitem[Packer et~al.(2023)Packer, Wooders, Lin, Fang, Patil, Stoica, and
  Gonzalez]{packer2023memgpt}
Charles Packer, Sarah Wooders, Kevin Lin, Vivian Fang, Shishir~G. Patil, Ion
  Stoica, and Joseph~E. Gonzalez.
\newblock {MemGPT}: Towards {LLM}s as operating systems.
\newblock \emph{arXiv preprint arXiv:2310.08560}, 2023.

\bibitem[Park et~al.(2023)Park, O'Brien, Cai, Morris, Liang, and
  Bernstein]{park2023generative}
Joon~Sung Park, Joseph~C. O'Brien, Carrie~J. Cai, Meredith~Ringel Morris, Percy
  Liang, and Michael~S. Bernstein.
\newblock Generative agents: Interactive simulacra of human behavior.
\newblock In \emph{Proceedings of the 36th Annual ACM Symposium on User
  Interface Software and Technology (UIST)}, 2023.

\bibitem[Reimers and Gurevych(2019)]{reimers2019sbert}
Nils Reimers and Iryna Gurevych.
\newblock Sentence-{BERT}: Sentence embeddings using {S}iamese {BERT}-networks.
\newblock In \emph{Proceedings of EMNLP-IJCNLP}, pages 3982--3992, 2019.

\bibitem[Rogers et~al.(1977)Rogers, Kuiper, and Kirker]{rogers1977sre}
T.~B. Rogers, N.~A. Kuiper, and W.~S. Kirker.
\newblock Self-reference and the encoding of personal information.
\newblock \emph{Journal of Personality and Social Psychology}, 35\penalty0
  (9):\penalty0 677--688, 1977.

\bibitem[Sumers et~al.(2024)Sumers, Yao, Narasimhan, and
  Griffiths]{sumers2024cogarch}
Theodore~R. Sumers, Shunyu Yao, Karthik Narasimhan, and Thomas~L. Griffiths.
\newblock Cognitive architectures for language agents.
\newblock \emph{Transactions on Machine Learning Research}, 2024.

\bibitem[Sutton and Barto(2018)]{sutton2018rl}
Richard~S. Sutton and Andrew~G. Barto.
\newblock \emph{Reinforcement Learning: An Introduction}.
\newblock MIT Press, 2nd edition, 2018.

\bibitem[Wu et~al.(2025)Wu, Wang, Yu, Zhang, Chang, and Yu]{wu2025longmemeval}
Di~Wu, Hongwei Wang, Wenhao Yu, Yuwei Zhang, Kai-Wei Chang, and Dong Yu.
\newblock {LongMemEval}: Benchmarking chat assistants on long-term interactive
  memory.
\newblock In \emph{International Conference on Learning Representations
  (ICLR)}, 2025.
\newblock Metadata to be re-verified at submission.

\bibitem[Zhong et~al.(2024)Zhong, Guo, Gao, Ye, and Wang]{zhong2024memorybank}
Wanjun Zhong, Lianghong Guo, Qiqi Gao, He~Ye, and Yanlin Wang.
\newblock {MemoryBank}: Enhancing large language models with long-term memory.
\newblock \emph{Proceedings of the AAAI Conference on Artificial Intelligence},
  38\penalty0 (17):\penalty0 19724--19731, 2024.

\end{thebibliography}

\end{document}